\documentclass[10pt,twocolumn,letterpaper]{article}

\usepackage[pagenumbers]{cvpr} 

\usepackage{graphicx}
\usepackage{amsmath}
\usepackage{amssymb}
\usepackage{booktabs}
\usepackage{float}

\usepackage[pagebackref,breaklinks,colorlinks]{hyperref}

\usepackage[capitalize]{cleveref}
\crefname{section}{Sec.}{Secs.}
\Crefname{section}{Section}{Sections}
\Crefname{table}{Table}{Tables}
\crefname{table}{Tab.}{Tabs.}

\def\cvprPaperID{7843} 
\def\confName{CVPR}
\def\confYear{2023}

\begin{document}

\title{Next3D: Generative Neural Texture Rasterization for 3D-Aware Head Avatars}

\author{
Jingxiang Sun$^{1}$ \qquad Xuan Wang$^{2}$ \qquad Lizhen Wang$^{1}$ \qquad Xiaoyu Li$^{3}$ \qquad Yong Zhang$^{3}$ \vspace{2pt} \\
Hongwen Zhang$^{1}$ \qquad Yebin Liu$^{1}$ \vspace{5pt} \\
$^{1}$Tsinghua University \qquad
$^{2}$Ant Group \qquad
$^{3}$Tencent AI Lab \qquad
}

\twocolumn[{%
\renewcommand\twocolumn[1][]{#1}%

\maketitle

\begin{center}
    \centering
    \includegraphics[width=\textwidth]{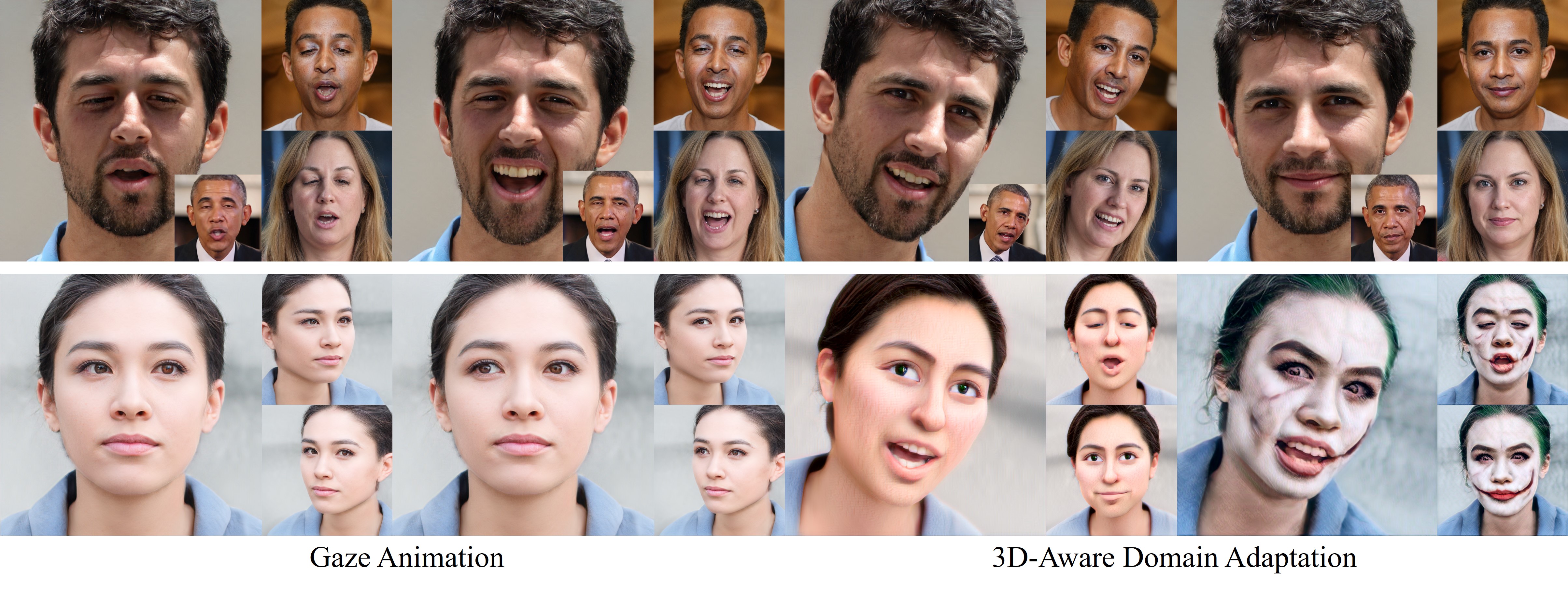}
    \captionof{figure}{Our 3D GAN synthesizes generative, high-quality, and 3D-consistent facial avatars from unstructured 2D images. Unlike current animatable 3D GANs that only modify yaw-pitch head poses and facial expressions, our approach enables fine-grained control over full-head rotations, facial expressions, eye blinks, and gaze directions with strict 3D consistency and a high level of photorealism. Our approach also provides strong 3D priors for downstream tasks such as 3D-aware stylization.}
\end{center}%
}]


\begin{abstract}
3D-aware generative adversarial networks (GANs) synthesize high-fidelity and multi-view-consistent facial images using only collections of single-view 2D imagery. Towards fine-grained control over facial attributes, recent efforts incorporate 3D Morphable Face Model (3DMM) to describe deformation in generative radiance fields either explicitly or implicitly. Explicit methods provide fine-grained expression control but cannot handle topological changes caused by hair and accessories, while implicit ones can model varied topologies but have limited generalization caused by the unconstrained deformation fields. We propose a novel 3D GAN framework for unsupervised learning of generative, high-quality and 3D-consistent facial avatars from unstructured 2D images. To achieve both deformation accuracy and topological flexibility, we propose a 3D representation called Generative Texture-Rasterized Tri-planes. The proposed representation learns Generative Neural Textures on top of parametric mesh templates and then projects them into three orthogonal-viewed feature planes through rasterization, forming a tri-plane feature representation for volume rendering. In this way, we combine both fine-grained expression control of mesh-guided explicit deformation and the flexibility of implicit volumetric representation. We further propose specific modules for modeling mouth interior which is not taken into account by 3DMM. Our method demonstrates state-of-the-art 3D-aware synthesis quality and animation ability through extensive experiments. Furthermore, serving as 3D prior, our animatable 3D representation boosts multiple applications including one-shot facial avatars and 3D-aware stylization. 
\end{abstract}
\section{Introduction}

Animatable portrait synthesis is essential for movie postproduction, visual effects, augmented reality (AR), and virtual reality (VR) telepresence applications. Efficient animatable portrait generators should be capable of synthesizing diverse high-fidelity portraits with full control of the rigid head pose, facial expressions and gaze directions at a fine-grained level. The main challenges of this task lie in how to model accurate deformation and preserve identity through animation in the generative setting, i.e. training with only unstructured corpus of 2D images.


Several 2D generative models perform image animation by incorporating the 3D Morphable Face Models (3DMM)~\cite{10.1145/311535.311556} into the portrait synthesis~\cite{tewari2020stylerig, deng2020disentangled, doukas2021headgan, kim2018deep, thies2016face2face, xu2020deep, ren2021pirenderer, yin2022styleheat}. These 2D-based methods achieve photorealism but suffer from shape distortion during large motion due to a lack of geometry constraints. Towards better view consistency, many recent efforts incorporate 3DMM with 3D GANs, learning to synthesize animatable and 3D consistent portraits from only 2D image collections in an unsupervised manner~\cite{bergman2022gnarf, noguchi2022unsupervised, tang2022explicitly, yue2022anifacegan, hong2022eva3d, zhang2022avatargen, lin20223d, sun2022controllable}. Bergman et al. \cite{bergman2022gnarf} propose an explicit surface-driven deformation field for warping radiance fields. While modeling accurate facial deformation, it cannot handle topological changes caused by non-facial components, e.g. hair, glasses, and other accessories. AnifaceGAN~\cite{yue2022anifacegan} builds an implicit 3DMM-conditioned deformation field and constrains animation accuracy by imitation learning. It achieves smooth animation on interpolated expressions, however, struggles to generate reasonable extrapolation due to the under-constrained deformation field. Therefore, The key challenge of this task is modeling deformation in the 3D generative setting for animation accuracy and topological flexibility.




In this paper, we propose a novel 3D GAN framework for unsupervised learning of generative, high-quality, and 3D-consistent facial avatars from unstructured 2D images. Our model splits the whole head into dynamic and static parts, and models them respectively. For dynamic parts, the key insight is to combine both fine-grained expression control of mesh-guided explicit deformation and flexibility of implicit volumetric representation. To this end, we propose a novel representation, \textit{Generative Texture-Rasterized Tri-planes}, which learns the facial deformation through \textit{Generative Neural Textures} on top of a parametric template mesh and samples them into three orthogonal-viewed and axis-aligned feature planes through standard rasterization, forming a tri-plane feature representation. Such texture-rasterized tri-planes re-form high-dimensional dynamic surface features in a volumetric representation for efficient volume rendering and thus inherit both the accurate control of the mesh-driven deformation and the expressiveness of volumetric representations. Furthermore, we represent static components (body, hair, background, etc.) by another tri-plane branch, and integrate both through alpha blending.

Another key insight of our method is to model the mouth interior which is not taken into account by 3DMM. Mouth interior is crucial for animation quality but often ignored by prior arts. We propose an efficient teeth synthesis module, formed as a style-modulated UNet, to complete the inner mouth features missed by the template mesh. To further regularize the deformation accuracy, we introduce a deformation-aware discriminator which takes as input synthetic renderings, encouraging the alignment of the final outputs with the 2D projection of the expected deformation. 


To summarize, the contributions of our approach are:

\begin{itemize}
    \item We present an animatable 3D-aware GAN framework for photorealistic portrait synthesis with fine-grained animation, including expressions, eye blinks, gaze direction and full head poses.  
    
    \item We propose \textit{Generative Texture-Rasterized Triplanes}, an efficient deformable 3D representation that inherits both fine-grained expression control of mesh-guided explicit deformation and flexibility of implicit volumetric representation. To our knowledge, we are the first method to incorporate Neural Textures into animatable 3D-aware synthesis.  
    \item Our learned generative animatable 3D representation can serve as a strong 3D prior and boost the downstream application of 3D-aware one-shot facial avatars. Our model also pushes the frontier of 3D stylization with high-quality out-of-domain facial avatars.
\end{itemize}
\section{Related Work}
\begin{figure*}[htbp]
  \centering
  \includegraphics[width=\textwidth]{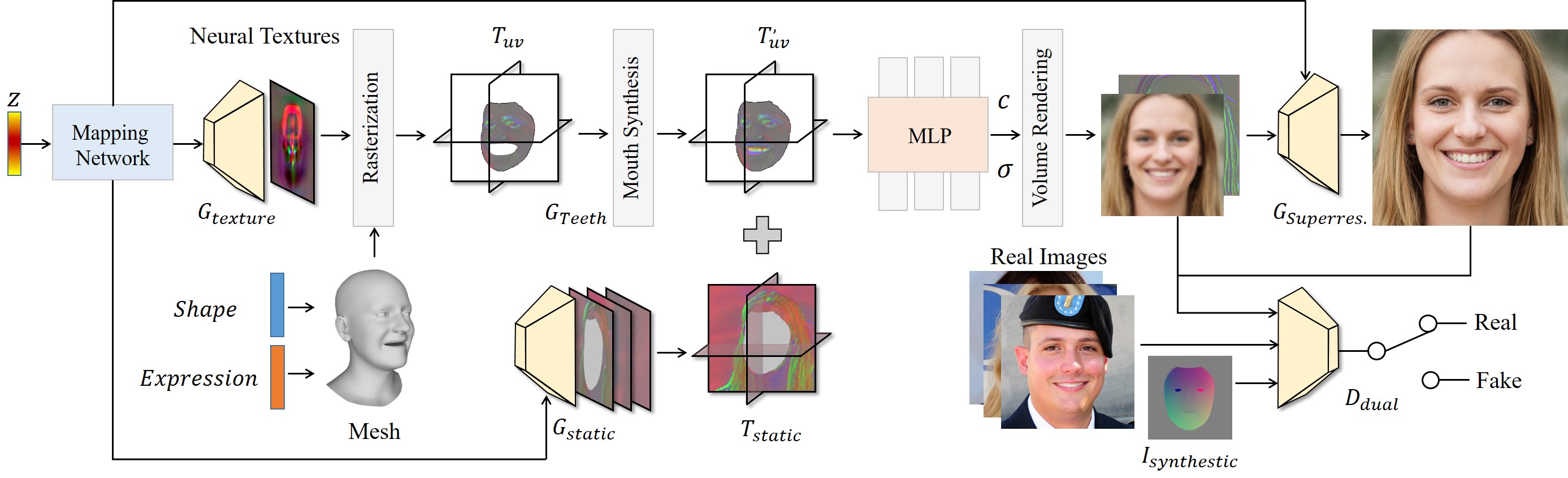}
  \caption{Our 3D GAN framework consists of two tri-plane branches $T_{uv}$ and $T_{static}$ modeling dynamic and static components. $T_{uv}$ is formed by the orthogonal rasterized Generative Neural Textures which are synthesized by a StyleGAN generator, $G_{texture}$, on top of deformable template mesh. $T_{static}$ is synthesized by another StyleGAN generator, $G_{static}$. The mouth synthesis module, $G_{teeth}$, is presented for completing mouth interior. Blended triplanes are incorporated with hybrid neural renderer consisting of volume rendering and a super-resolution module $G_{superres.}$. For discrimination, synthetic renderings $I_{synthetic}$ is taken into the dual discriminator $D_{dual}$. }
  \label{pipeline}
\end{figure*}

\noindent\textbf{Generative 3D-aware Image Synthesis.} Generative adversarial networks~\cite{goodfellow2014generative} have achieved photorealistic synthesis in 2D domain. Building on the success of 2D GANs, many efforts have lifted the image synthesis into 3D with explicit view control. Early voxel-based approaches~\cite{gadelha20173d,henzler2019escaping,nguyen2019hologan,nguyen2020blockgan,wu2016learning,zhu2018visual} adopt the 3D CNN generators whose heavy computational burden limits the high-resolution image synthesis. Recent works incorporate more efficient neural scene representations, such as fully implicit networks~\cite{chan2021pi,schwarz2020graf,deng2021gram,zhou2021cips3d, chen2022gdna, epigraf, Chen2022Sem2NeRFCS, pan2021shading, cai2022pix2nerf}, sparse voxel grids~\cite{schwarz2022voxgraf}, multiple planes~\cite{zhao-gmpi2022} or a combination of low-resolution feature volume and 2D super-resolution~\cite{xue2022giraffe, niemeyer2021giraffe, gu2021stylenerf, orel2021stylesdf, eg3d, xu20223d, Zhang20223DAwareSG, zhang2022multi}. We leverage the tri-plane representation proposed in \cite{eg3d} and endow it with animation ability by the orthogonal-rasterized \textit{Generative Neural Textures}. 


Some other current works~\cite{sun2022fenerf,sun2022ide,kwak2022injecting,noguchi2022unsupervised,bergman2022gnarf,tang2022explicitly,yue2022anifacegan,hong2022eva3d, zhuang2022controllable, zhang2022training} focus on the editability of 3D-aware generative models. FENeRF~\cite{sun2022fenerf} and IDE-3D~\cite{sun2022ide} perform semantic-guided 3D face editing by incorporating semantic-aware radiance fields and GAN inversion. However, they cannot produce continuous and stable editing on videos. Other methods~\cite{noguchi2022unsupervised,bergman2022gnarf,tang2022explicitly,yue2022anifacegan,hong2022eva3d} employ 3D priors to achieve animatable image synthesis and the main differences lie on the deformation strategies including linear blend skinning~\cite{noguchi2022unsupervised,hong2022eva3d}, surface-driven deformation~\cite{bergman2022gnarf}, 3DMM-guided latent decomposition~\cite{tang2022explicitly} and neural deformation fields~\cite{yue2022anifacegan}. 
These approaches either don't allow for topology changes or need elaborate loss design to ensure the accuracy of deformation. On the contrary, our approach naturally achieves accurate animation with explicit mesh guidance and further allow for topology changes by adapting surface deformation into a continuous volumetric representation.


\noindent\textbf{Facial animation with 3D morphable face models.} Blanz et al.~\cite{10.1145/311535.311556} model facial texture and shape as vector spaces, known as the 3D Morphable Model (3DMM). Extensions of 3DMM, such as full-head PCA models~\cite{dai2020statistical, ploumpis2020towards}, blend-shape models~\cite{li2017learning}, are extensively studied and widely used in facial animation tasks~\cite{gecer2019ganfit, thies2016face2face}. Benefiting from 3DMM, these methods can model the deformation of facial parts accurately and continuously, nevertheless, struggle to represent non-facial areas missed by 3DMM, e.g. hair, teeth, eyes, and body. Moreover, these methods are prone to lacking facial details. To fill the missing areas and complete more realistic facial details, later works~\cite{kim2018deep, xu2020deep, thies2019deferred, gecer2018semi, fried2019text, tewari2020stylerig, deng2020disentangled, doukas2021headgan, ren2021pirenderer} apply learned approaches on top of 3DMM renderings. DiscoFaceGAN~\cite{deng2020disentangled} maps 3DMM parameters into the latent space and decouple them by imitative-contrastive learning. Though efficient and photorealistic animation is achieved, these 2D methods don't model 3D geometry and thus cannot remain strict 3D consistency with large head pose changes. For strict 3D consistency, recent efforts~\cite{gafni2021dynamic, zheng2022avatar, grassal2022neural, zhang2022fdnerf, athar2022rignerf} incorporate 3DMM into volumetric representations~\cite{mildenhall2020nerf, sitzmann2019deepvoxels, kanazawa2018learning, peng2020convolutional, hong2022headnerf} to achieve view-consistent facial animation. Furthermore, volumetric representations enable the modeling of thin structures such as hair, and also mouth interior thanks to its spatial continuity. While 3DMM have been adopted to animate radiance fields for single-scene scenarios, it is challenging to adapt to the generative setting with the absence of groundtruth supervision.

\noindent\textbf{Neural scene representations.} The neural scene representations can be roughly categorized into implicit and explicit surface representations and volumetric representations~\cite{tewari2022advances}. The surface can be represented explicitly by point clouds\cite{pfister2000surfels,lange2021fc2t2}, meshes~\cite{burov2021dynamic, thies2019deferred, oechsle2019texture, baatz2021nerf}, or defined implicitly as a zero level-set of a function~\cite{park2019deepsdf,chen2019learning,xu2019disn}, like signed distance function, which can be approximated by coordinate-based multi-layer perceptrons (MLPs). On the contrary, volumetric representations~\cite{mildenhall2020nerf, sitzmann2019deepvoxels, kanazawa2018learning, peng2020convolutional} store volumetric properties (occupancies, radiance, colors, etc.) instead of the surface of an object. These properties can be stored in explicit voxel grids~\cite{sitzmann2019deepvoxels,peng2020convolutional,kanazawa2018learning} or the weights of a neural network implicitly~\cite{mildenhall2020nerf}. In this work, we propose a hybrid surface–volumetric representation. Specifically, we learn the deformable surface radiance by neural textures~\cite{thies2019deferred, grigorev2021stylepeople} on top of the template mesh and rasterize it into three orthogonal-viewed feature planes. Then, the planes are reshaped to a tri-plane representation with decoding into neural radiance fields for volume rendering.

\section{Approach}

We present an animatable 3D-aware facial generator that equips with fine-grained expression and pose control, photorealistic rendering quality, and high-quality underlying geometry. The proposed method models dynamic and static components by two independent tri-plane branches. Specifically, we propose \textit{Generative Texture-rasterized Tri-planes} for modeling dynamic facial parts (Sec.~\ref{sec:3.1}). Furthermore, we propose an efficient mouth synthesis module to complete the mouth interior that is not included in 3DMM (Sec.~\ref{sec:3.2}). We further adopt another tri-plane branch for the static components (Sec.~\ref{sec:3.3}). Both tri-planes are blended together for hybrid neural rendering (Sec.~\ref{sec:3.4}). We introduce an deformation-aware discriminator (Sec.~\ref{sec:3.5}) and illustrate the training objectives in Sec.~\ref{sec:3.6}.

\subsection{Generative texture-rasterized tri-planes}
\label{sec:3.1}
EG3D~\cite{eg3d} presents an efficient tri-plane-based hybrid 3D representation to synthesize high-resolution images with multi-view consistency. Nonetheless, EG3D lacks control over facial deformations and thus cannot be directly applied to animation tasks. To this end, we leverage Neural textures~\cite{thies2019deferred} to represent deformable facial parts. In general, Neural Textures are a set of learned high-dimensional feature maps that can be interpreted by a neural renderer. We extend it to our generative setting and synthesize the neural textures through a StyleGAN2 CNN generator $G_{texture}$. As shown in Fig.~\ref{pipeline}, we first sample a latent code $z$ and map it into an intermediate latent space by the mapping network. Our texture generator architecture closely follows StyleGAN2 backbone \cite{karras2020analyzing}, except producing a $256 \times 256 \times 32$ neural texture map, $T$, instead of a three-channel RGB image. Storing a high-dimensional learned feature vector per texel, $T$ can be rasterized to a view-dependent screen-space feature map given a mesh with uv-texture parameterization and a target view as input. In our case, we use the FLAME template~\cite{li2017learning} to provide a coarse mesh that can be driven by deformation parameters. Given the pre-designed texture mapping function, we employ the standard graphics pipeline to rasterize the neural textures from the texture space into the screen space based on the template mesh. We choose \textit{Neural Textures} as the deformation method for two reasons. First, compared with other explicit deformation (e.g. linear blend skinning and surface-driven deformation) highly dependent on the accurate underlying geometry, Neural Textures embed high-level features which compensate the imperfect geometry and thus are more suitable for our settings where template meshes are not accurate. Furthermore, unlike implicit deformation methods~\cite{yue2022anifacegan, deng2020disentangled, tang2022explicitly}, our explicit mesh-guided deformation alleviates the requirement of elaborate imitation learning while gain better expression generalization (Fig.~\ref{comparison}). 

Neural Textures encode surface deformation accurately with mesh guidance but lack generalization to 3D points far from surface. Besides, it also doesn't allow for topological changes. To this end, we propose Generative Texture-Rasterized Tri-planes, $T_{uv}$ which reshapes the rasterized textures into a tri-plane representation. Therefore, we can adapt such surface deformation into a continuous volume. Specifically, we rasterize neural textures based on the template mesh into three orthogonal views and place them in three axis-aligned feature planes. In practice, considering the zygomorphy, the rasterization is applied at both the left and right views and the rasterized features are concatenated for one single plane by summation. In this way, Our hybrid surface-volumetric representation inherits the best of both worlds: accurate mesh-guided facial deformation of Neural Textures and the continuity and topological flexibility of volumetric representations. See Fig.~\ref{geometry} for the topological-aware animation of portraits with glasses. 

\subsection{Mouth synthesis module}
\label{sec:3.2}

\begin{figure}[htbp]
  \centering
  \includegraphics[width=\linewidth]{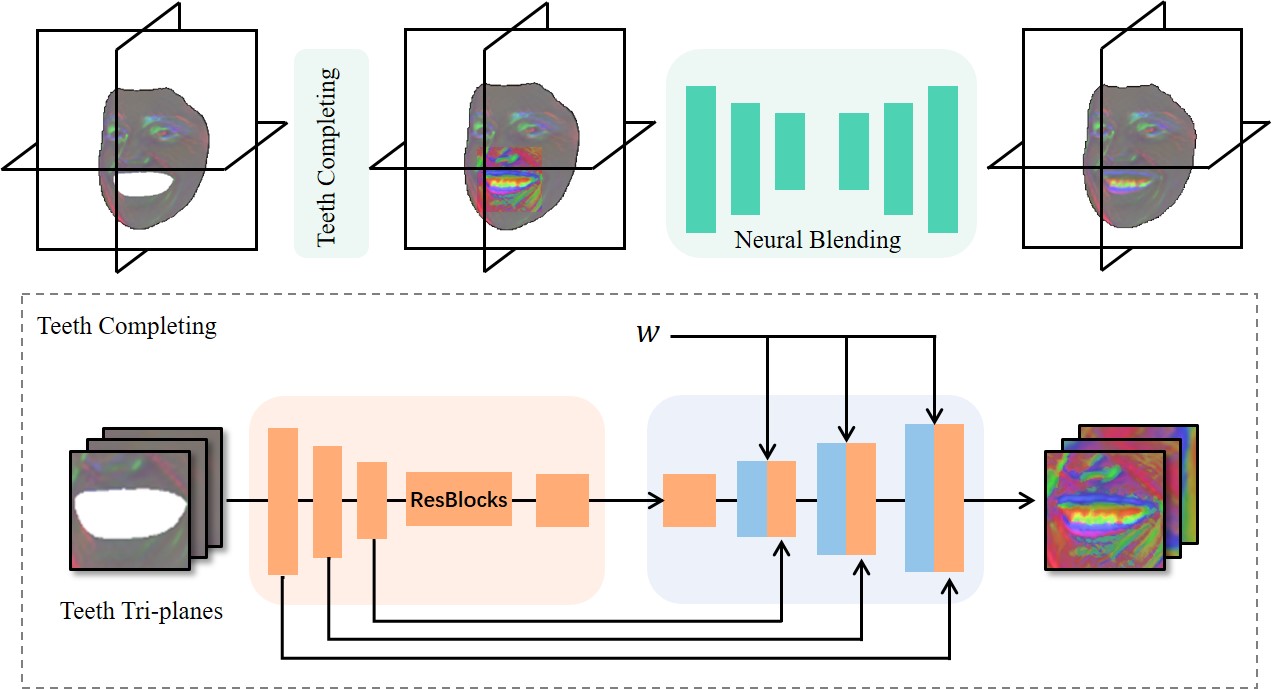}
  \caption{The teeth synthesis module consists of a teeth completing module and a neural blending module. The teeth competing module produces mouth interior conditioned on multi-scale teeth exterior features and latent codes $w$. }
  \label{fig:teeth}
\end{figure}

Since the FLAME template doesn't contain inner mouth, we propose a teeth synthesis module, $G_{teeth}$, to complete the missing teeth features in $T_{uv}$. As shown in Fig.~\ref{fig:teeth}, for each feature plane of $T_{uv}$, we crop the teeth area by the expanded landmarks and resize it into $64 \times 64$. Then, the stacked mouth features are processed by $G_{teeth}$ which employs a style-modulated UNet~\cite{wang2022faceverse}. The downsampling process of $G_{teeth}$ encodes $f_{teeth}$ into multi-scale feature maps which serve as content conditions for the following StyleGAN layers. The output teeth features $f^{'}_{teeth}$ are transformed inversely and concatenated with the feature planes of $T_{uv}$. To eliminate the texture flickering of mouth boundary, we further feed $T_{uv}$ into a shallow-UNet-based neural blending module and obtain $T_{uv}^{'}$. We conduct a series of ablation studies and prove that the proposed teeth synthesis module brings a remarkable improvement on both the animation accuracy and synthesis quality (Sec.~\ref{sec:4.2}). 

\subsection{Modeling static components}
\label{sec:3.3}
The generative texture-rasterized tri-planes manage to model dynamic faces varying expressions and shapes, though, it is challenging to synthesize static parts like diverse haircut, background and upper body which are not included in the FLAME template. To this end, we model these parts by another tri-plane branch, $T_{static}$, which is generated by a StyleGAN2 CNN generator $G_{static}$ sharing the same latent code with $G_{texture}$. The plane features of $T_{uv}^{'}$ and $T_{static}$ are blended on each plane by the alpha masks rendered by rasterization. Such design not only benefits the modeling of various-styled static components but also enforces their consistency during facial animations.       

\subsection{Neural rendering}
\label{sec:3.4}
Given the blended tri-planes, for any point in the 3D space, we project it into each plane and sample the features bi-linearly. Then, the sampled features are aggregated by summation and decoded into volume density $\sigma$ and feature $f$ by a lightweight decoder. Similar to \cite{sun2022ide, eg3d}, the decoder is a single hidden layered multi-layer perceptron (MLP) with softplus activation. The volume rendering is employed to accumulate $\sigma$ and $f$ along the rays cast through each pixel to compute a 2D feature image $I_{f}$. Similar to \cite{eg3d, gu2021stylenerf, orel2021stylesdf}, we leverage a 2D super-resolution module $G_{superres.}$ to interpret the feature image into RGB image $I_{RGB}$ with higher resolution. The super-resolution module consists of three StyleGAN2 synthesis blocks and the noise input is removed for alleviating texture flickers. In our case, $I_{f}$ and $I_{RGB}$ are set to $64 \times 64$ and $512 \times 512$, respectively.

\subsection{Deformation-aware discriminator}
\label{sec:3.5}
To learn the unsupervised 3D representations, we adopt a 2D convolutional discriminator $D$ to critique the renderings. Inspired by \cite{eg3d}, we regularize the first three channels of $I_{f}$ as low-resolution RGB image, which is concatenated with $I_{RGB}$ as the input for the discriminator. However, the discrimination with only image input can only ensure that the deformed images are always in a correct distribution instead of matching the expected deformations. Therefore, we make the discriminator aware of the expression and shape under which the generated image are deformed. For the same purpose, GNARF~\cite{bergman2022gnarf} conditions the discriminator on the FLAME parameters by concatenating them as the input to the mapping network. However, we find empirically that such a conditioning method leads to training instability, consistent with \cite{bergman2022gnarf}. Instead, we re-render the template mesh under the rendered pose to get the synthetic rendering $I_{synthetic}$ and feed it into $D_{dual}$ along with image pairs. Here, we adopt correspondence images inspired by \cite{kim2018deep}. Such concatenation encourages the final output to align with the synthetic rendering and learn the expected deformation.

\subsection{Training objectives}
\label{sec:3.6}
During training, we use the non-saturating GAN loss with R1 regularization. Moreover, we adopt the density regularization proposed in EG3D~\cite{eg3d}. Therefore, the total learning objective is: \vspace{-1em}

\begin{equation}
\small
    \begin{split}
    \mathcal{L}_{D_{dual}, G} = 
    &\mathbb{E}_{z \sim p_{z}, \epsilon \sim p_{\epsilon}}[f(D_{dual}(G(z, \epsilon)))] + \\
    &\mathbb{E}_{I^{r} \sim {p_{I^{r}}}}[f(-D_{dual}(I^{r}) + \\
    &\lambda\|\mathbf{\bigtriangledown}D_{dual}(I^{r})\|^2],
    \end{split}
    \label{eq:generator loss1}
\end{equation}
\vspace{-2em}

\begin{equation}
\small
    \begin{split}
    \mathcal{L}_{density} = 
    &\mathbf{\sum}_{x_{s} \in \mathcal{S}}\|d(x_{s}) - d(x_{s}+\epsilon)\|_2,
    \end{split}
    \label{eq:generator loss2}
\end{equation}
\vspace{-1em}
\begin{equation}
\small
    \begin{split}
    \mathcal{L}_{total} = 
    &\mathcal{L}_{D_{dual}, G} + \lambda_{density} \mathcal{L}_{density},
    \end{split}
    \label{eq:generator loss3}
\end{equation}

\noindent where $I^{r}$ is the combination of real images, blurred real images, and the corresponding synthetic renderings, which are sampled from the training set with distribution $p_{I}$. We adopt many training hyperparameters from EG3D and StyleGAN2 (learning rates of generator and discriminator, batch size, R1 regularization, etc.). We train our model based on the pretrained model of EG3D~\cite{eg3d} and continue to train on 4 3090 GPUs for roughly 4 days. Please refer to the supplemental material for the implementation details.

\section{Experiments}

In this section, we first show qualitative and quantitative comparisons to state-of-the-art 2D / 3D animatable generative facial models (Sec.~\ref{sec:4.1}), and then discuss the conducted ablation studies of our design choices (Sec.~\ref{sec:4.2}). Furthermore, we show various applications combining our efficient 3D representation with GAN technologies such as GAN inversion and style transfer (Sec.~\ref{sec:4.3}).

\noindent\textbf{Datasets.} We train and test our methods on FFHQ~\cite{karras2019style}. We augment FFHQ with horizontal flips and use an off-the-shelf pose estimator~\cite{deng2019accurate} to label images with the approximated camera extrinsic parameters and constant intrinsics. To support full pose animation, in-plane (roll) rotation is also considered. Furthermore, we use DECA~\cite{feng2021learning} to estimate the FLAME parameters of facial identity $\beta \in \mathbb{R}^{100}$, jaw pose $\theta_{jaw} \in \mathbb{R}^{3}$ and expression $\psi \in \mathbb{R}^{50}$. Since DECA doesn't account for eyeball movement, we additionally adopt an efficient facial detector~\footnote{https://mediapipe.dev/} to detect 2D landmarks of irises and optimize eye poses $\theta_{eye} \in \mathbb{R}^{6}$ by minimizing the re-projection errors. Based on these FLAME parameters, we produce a template mesh with 5023 vertices and 9976 faces to drive facial deformations. 

\subsection{Comparisons}
\label{sec:4.1}

\begin{figure*}[htbp]
  \centering
  \includegraphics[width=\textwidth]{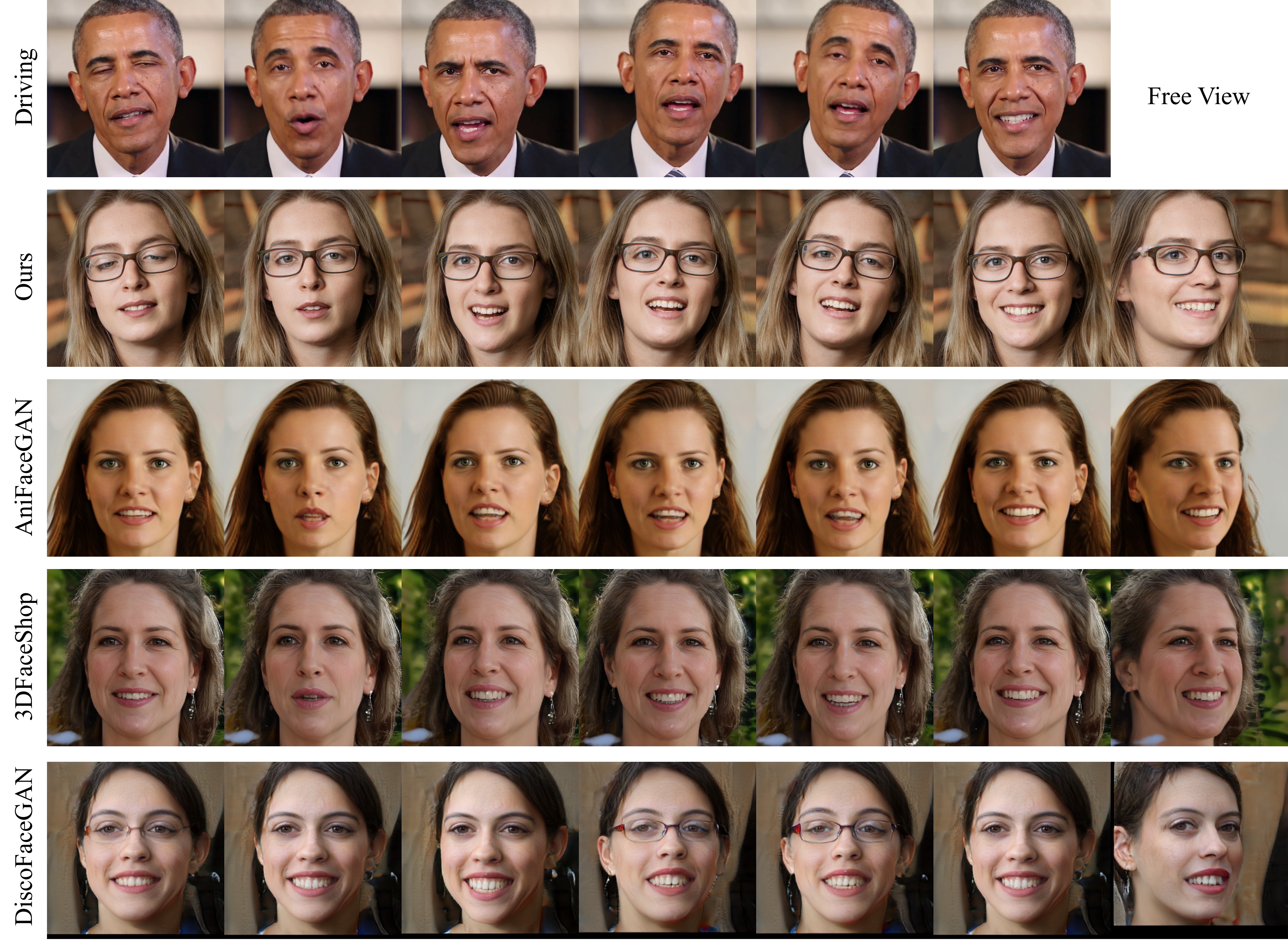}
  \caption{Comparison with the state-of-the-art animatable 3D \& 2D image synthesis methods. We extract several frames from a video clip and use the interpreted face model parameters to animate random virtual avatars.}
  \label{comparison}
\end{figure*}

\begin{figure}[htbp]
  \centering
  \includegraphics[width=\linewidth]{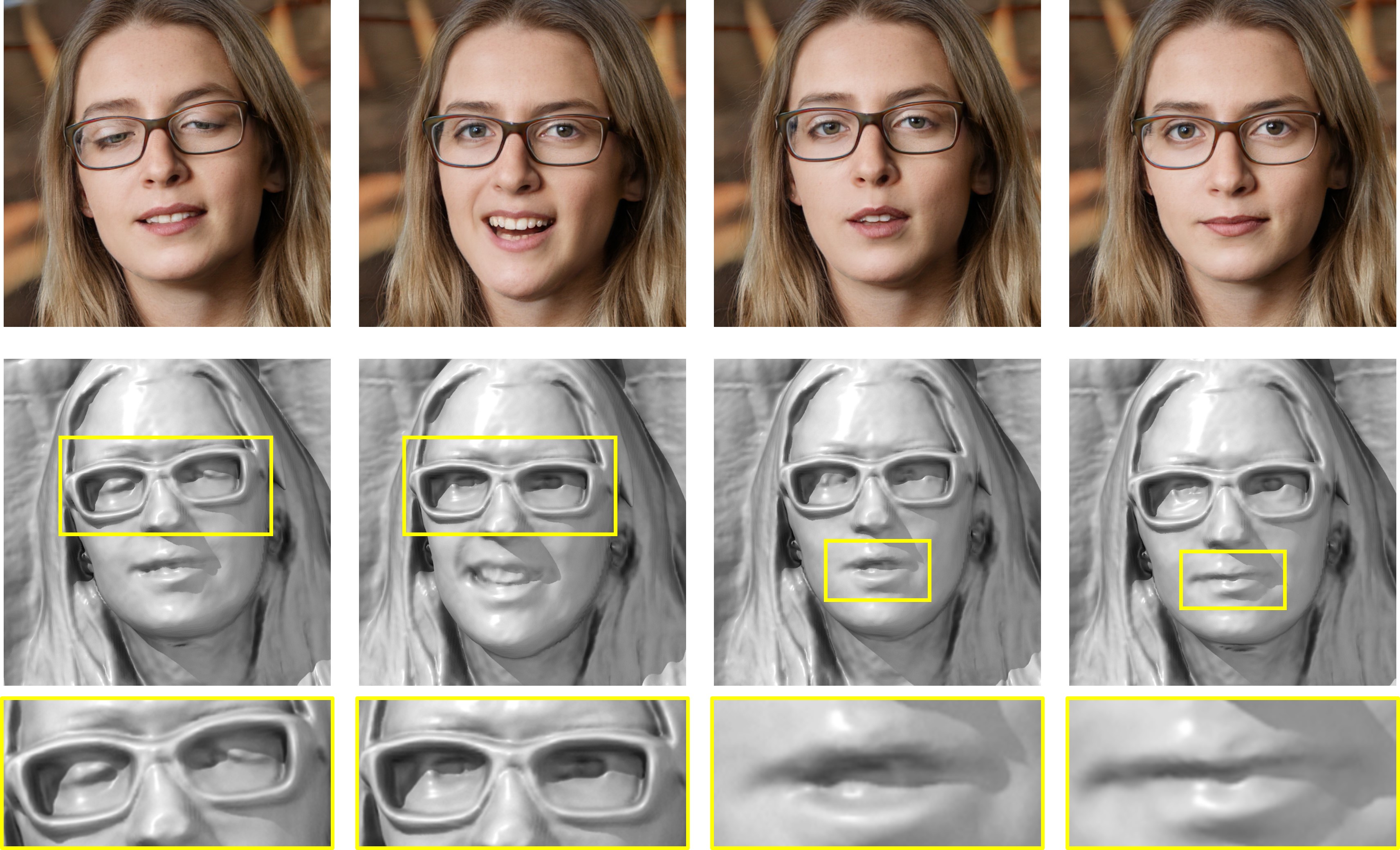}
  \caption{High-quality dynamic shapes with topological changes. As can be seen, we model detailed dynamic shapes of eyelids and lips, while keeping glasses unchanged.}
    \vspace{-1em}
  \label{geometry}
  
\end{figure}

\noindent\textbf{Baselines.} We compare our method against two state-of-the-art methods for animatable 3D-aware image synthesis: 3DFaceShop~\cite{tang2022explicitly}, and AniFaceGAN~\cite{yue2022anifacegan}. Besides, we also select DiscoFaceGAN~\cite{deng2020disentangled} as a baseline, which generates animatable 2D portraits conditioned on 3DMM parameters.

\noindent\textbf{Qualitative comparison.} Fig.~\ref{comparison} provides a qualitative comparison against baselines.
Overall, as can be seen, our method outperforms all baselines by some margin on both synthesis quality and animation accuracy. Specifically, DiscoFaceGAN~\cite{deng2020disentangled} suffers from inconsistent identity during animation. Moreover, it cannot generate reasonable mouth interior, e.g. stretched teeth. 3DFaceshop and AnifaceGAN synthesize 3D-consistent images, nevertheless, still struggle to model consistent mouth interior with the driving images. This is because their implicit deformation approaches are under-constrained, leading to overfit the expression bias (smiling with mouth half-opened) of datasets. Compared to the other methods, our approach not only synthesizes images with higher quality, but also preserves more detailed expressions of the driver images, including mouth interior, eye blinks and eye movements. Furthermore, we are the only method that supports in-plane head rotations. Fig.~\ref{geometry} provides visual examples of the synthesized high-quality geometry. Our approach can model detailed shape deformations (see zoomed eyelids and lips in Fig.~\ref{geometry}) with topological awareness, i.e. glasses are kept unchanged.

\begin{table}[t]
\scriptsize
\centering

\begin{tabular}{llllll} 
\toprule
    &\multicolumn{1}{c}{FID$\downarrow$} & \multicolumn{1}{c}{AED$\downarrow$} &  \multicolumn{1}{c}{APD$\downarrow$} & \multicolumn{1}{c}{APD*$\downarrow$} & \multicolumn{1}{c}{ID$\uparrow$} \\ 
\midrule  

DiscoFaceGAN~($256^2$)~\cite{deng2020disentangled} & \multicolumn{1}{c}{17.1} & 0.42 & 0.046 & 0.024 & 0.73 \\

AniFaceGAN~($256^2$)~\cite{yue2022anifacegan}  & \multicolumn{1}{c}{20.1} & 0.25 & 0.041 & 0.022 & 0.82 \\

3DFaceShop~($512^2$)~\cite{tang2022explicitly} & \multicolumn{1}{c}{23.7} & 0.31 & 0.045 & 0.024 & 0.75  \\

Ours~($512^2$) & \multicolumn{1}{c}{\textbf{3.9}} & \textbf{0.16} & \textbf{0.023} & \textbf{0.019} & \textbf{0.84} \\
\bottomrule
\end{tabular}
\caption{Quantitative comparison using FID,  average expression distance (AED), average pose distance (APD), and identity consistency (ID) for FFHQ. APD* means calculating pose distance with roll fixed.}
\label{tab:comparison}
\end{table}









\begin{figure}[htbp]
  \centering
  \includegraphics[width=\linewidth]{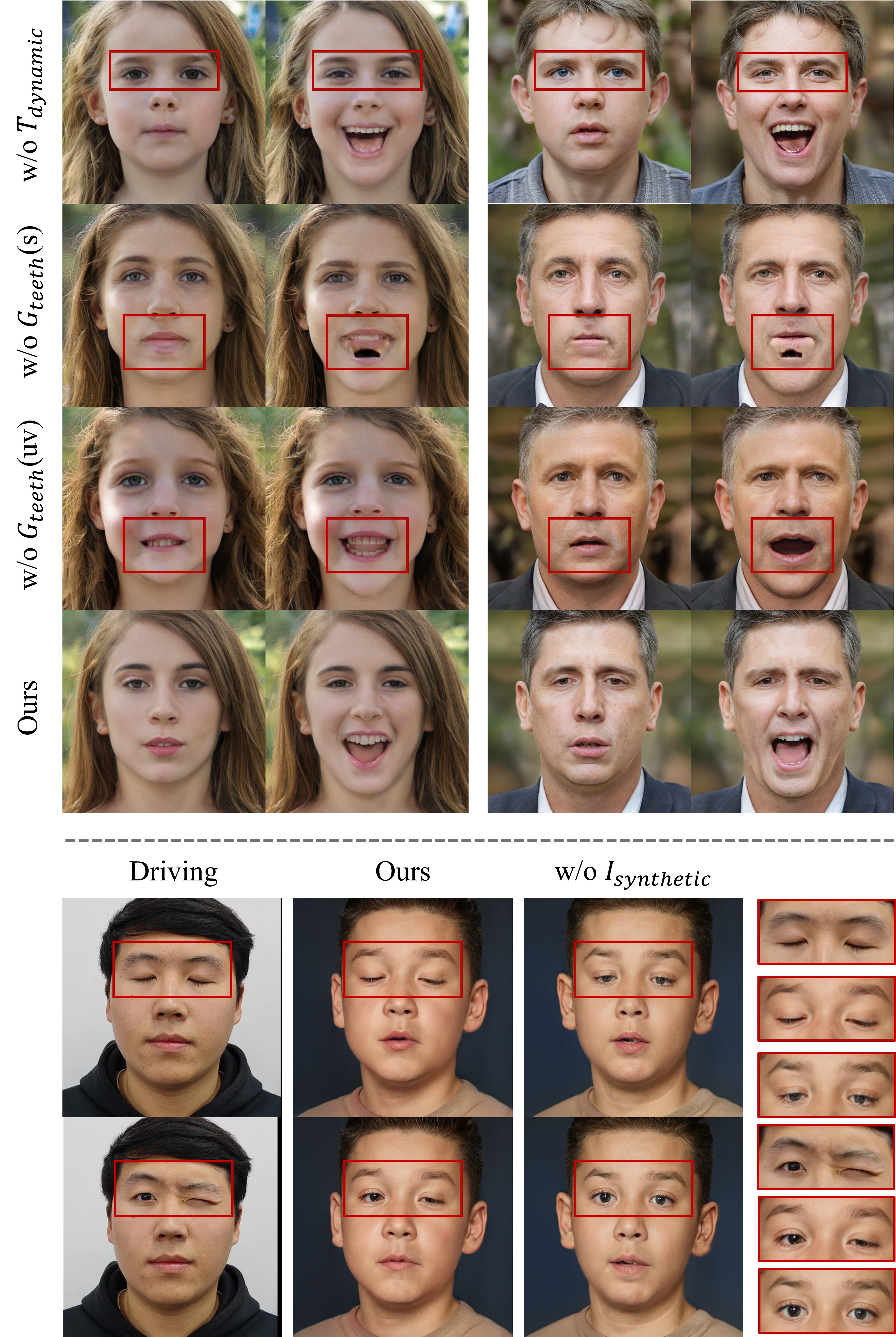}
  \caption{Ablation study on model designs. $T_{static}$ encourages better identity consistency; $G_{teeth}$ benefits realistic mouth interior; the discriminator with $I_{synthetic}$ input allows more consistent reconstruction of detailed expressions.}
  \label{ablation}
  \vspace{-1em}
\end{figure}

\noindent\textbf{Quantitative evaluation.} Tab.~\ref{tab:comparison} demonstrates quantitative results comparing our method against baselines evaluated on several metrics. We measure image quality with Frechet Inception Distance (FID)~\cite{heusel2017gans} between the entire FFHQ dataset and 50k generated images using randomly sampled latent codes, camera poses, and FLAME parameters. Since AniFaceGAN and DiscoFaceGAN synthesize images on the resolution of $256^2$, we test FID of them on FFHQ$256^2$. Following \cite{lin20223d, bergman2022gnarf}, we evaluate the faithfulness of the animation with the Average Expression Distance (AED), the Average Pose Distance (APD), and identity consistency (ID). For each method, we randomly sample 500 identities and animate each with randomly sampled 20 FLAME parameters of expressions and poses. Then, we estimate the FLAME parameters for these 10000 generated images and the average distances between the driving FLAME parameters and the reconstructed ones. For identity consistency, we randomly sample 2000 poses, 2000 sets of FLAME parameters, and 1000 identities. Then we randomly select two poses and two sets of FLAME parameters for each identity, generating a total of 1000 image pairs. We calculate consistency metric using a pre-trained Arcface model~\cite{deng2019accurate} for each image pair and report the average result. Since the other baselines don't support in-plane (roll) head rotation, we further report APD* which only accounts for poses on yaw and pitch. Our method achieves the best performance on all metrics. Note that our model demonstrates significant improvements in FID, bringing animatable 3D GAN to the same level as unconditional 3D GANs (4.7 for EG3D~\cite{eg3d}). For AED and APD, we also show superiority against baselines. Note that we still achieve the best pose consistency (0.019) when only considering yaw and pitch.

\subsection{Ablation study}


\begin{table}[t]
\label{table: ablation}
\small
\centering

\begin{tabular}{lllll} 
\toprule
    &\multicolumn{1}{c}{FID$\downarrow$} & \multicolumn{1}{c}{AED$\downarrow$} & \multicolumn{1}{c}{APD$\downarrow$} & \multicolumn{1}{c}{ID$\uparrow$} \\ 
\midrule  
w/o $T_{static}$ & \multicolumn{1}{c}{6.6} & 0.25 & 0.026 & 0.63 \\

w/o $G_{teeth} (uv)$ & \multicolumn{1}{c}{7.2} & 0.37 & 0.042 & 0.71 \\

w/o $G_{teeth} (s)$  & \multicolumn{1}{c}{8.4} & 0.32 & 0.036 & 0.79 \\

w/o $I_{synthetic}$  & \multicolumn{1}{c}{\textbf{3.8}} & 0.18 & 0.025 & 0.74 \\

Ours & \multicolumn{1}{c}{3.9} & \textbf{0.16} & \textbf{0.023} & \textbf{0.84} \\
\bottomrule
\end{tabular}
\caption{Ablation study on model designs. Modeling static components by $T_{static}$ improves identity consistency. The teeth synthesis $G_{teeth}$ module benefits both animation and synthesis quality significantly while adding synthetic renderings $I_{synthetic}$ into discrimination takes both a step further.}
\label{tab:ablation}
\end{table}

\label{sec:4.2}
\begin{figure*}[htbp]
  \centering
  \includegraphics[width=0.95\linewidth]{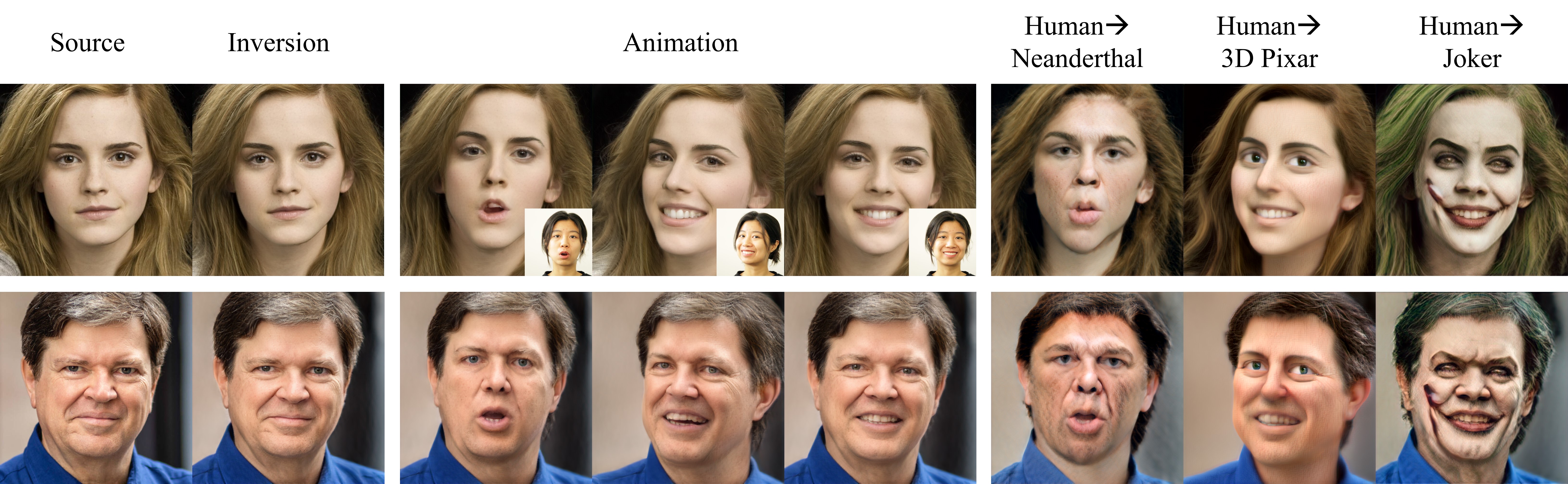}
  \caption{Applications of our model. We use PTI~\cite{roich2021pivotal} to fit 3D-aware avatars for real portraits and animate them with sampled video clips. Furthermore, we leverage StyleGAN-NADA~\cite{gal2022stylegan} into 3D settings and adapt these avatars into textually-prescribed domains.}
  \label{inversion}
\end{figure*}  

\begin{figure}[htbp]
  \centering
  \includegraphics[width=0.9\linewidth]{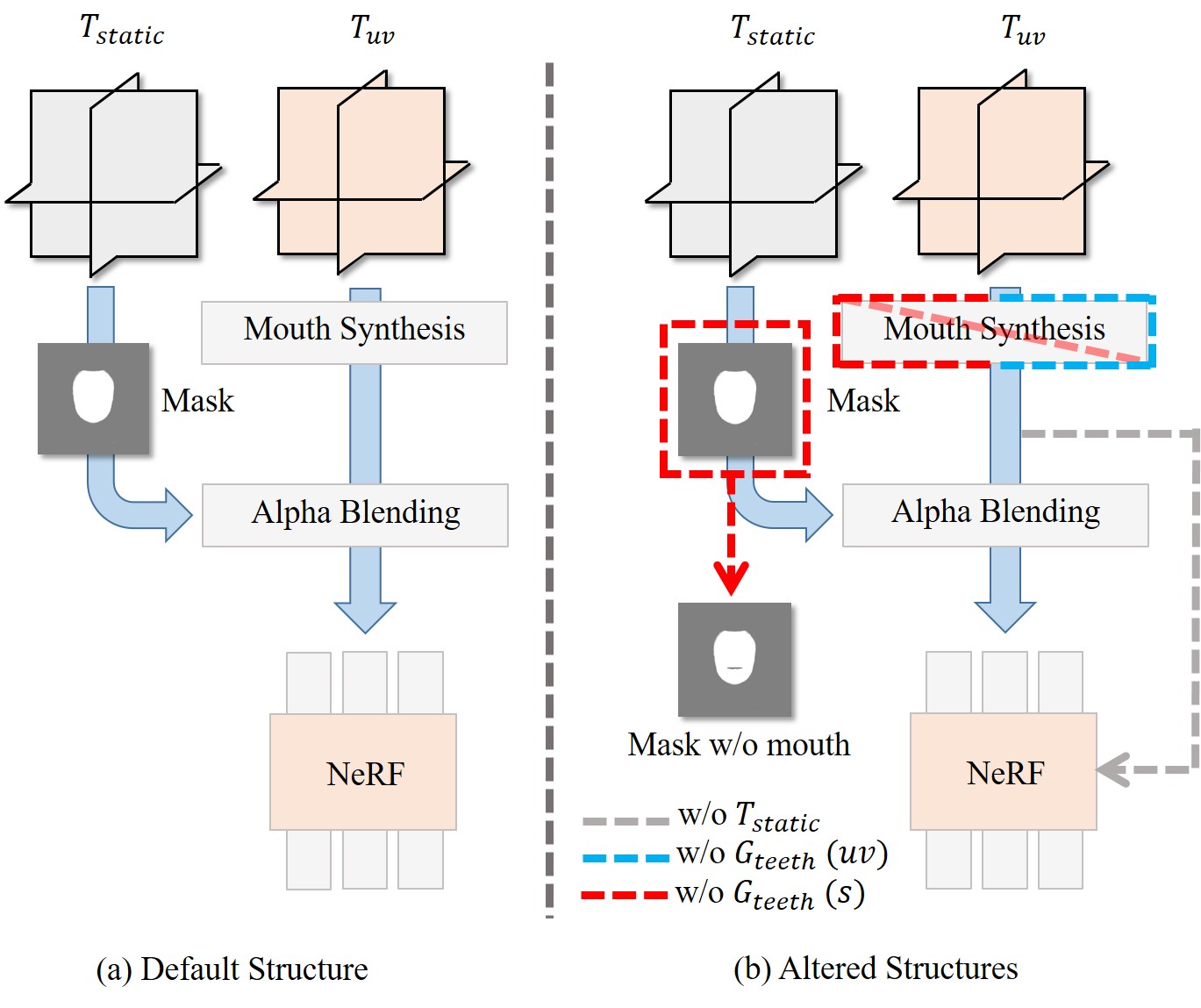}
  \caption{Illustrations of the proposed three variants of our model for ablation study.}
  \label{ablation_pipeline}
\end{figure}



\noindent\textbf{Static tri-planes.} As suggested by the grey lines in Fig.~\ref{ablation_pipeline}, this baseline removes the static tri-planes $T_{static}$ and entangles both dynamic and static components in $T_{uv}$. As illustrated in the first row of Fig.~\ref{ablation}, the identities change when varying expressions. Since there are no explicit constraints for identity consistency, the model would be prone to unexpected entanglement between expression and identity. Tab.~\ref{tab:ablation} shows a similar trend where removing $T_{static}$ leads to worse identity consistency.
 
\noindent\textbf{Mouth Synthesis.} When removing the mouth synthesis module $G_{teeth}$, we consider two altered choices: representing mouth features by $T_{static}$ or $T_{uv}$, named w/o $G_{teeth} (s)$ (red lines in Fig.~\ref{ablation_pipeline}) and w/o $G_{teeth} (uv)$ (blue lines in Fig.~\ref{ablation_pipeline}), respectively. The first baseline, illustrated in the second row of Fig.~\ref{ablation}, suffers from 'hole' artifacts of mouth. This is because inferior teeth move along with the jaw rotations and thus cannot be modeled by static features. The second baseline modeling teeth with $T_{uv}$ also leads to an unreasonable mouth interior since the FLAME template doesn't account for teeth area and the neural textures for teeth would be sampled from other unrelated areas leading to artifacts. Quantitatively, both baselines without $G_{teeth}$ show significant degradation in AED and APD. 

\noindent\textbf{Deformation-aware discriminator.} Tab.~\ref{tab:ablation} demonstrates that the deformation-aware discriminator with $I_{synthetic}$ input improves both animation accuracy and identity consistency, at the negligible expense of slightly reduced image quality. In Fig.~\ref{ablation}, we see that this design shows a better exploration of rare expressions, e.g. eye blinks.

\subsection{Applications}
\label{sec:4.3}

\noindent\textbf{One-shot portrait animation.} Fig.~\ref{inversion} shows the application of our model for one-shot head avatars. The learned generative animatbale 3D representation with expressive latent space can serve as a strong 3D prior for high-fidelity single-view 3D reconstruction and animation. Note that we can generate natural and consistent animations without video data training. 

\noindent\textbf{Animatable 3D-aware stylization.} Inspired by IDE-3D~\cite{sun2022ide}, we incorporate 2D CLIP-guided style tranfer methods~\cite{gal2022stylegan} with our animatable 3D representation for 3D-aware portrait stylization. The right three columns of Fig.~\ref{inversion} show examples of text-driven, stylized portrait animation. Specifically, to adapt a pre-trained model through only a textual prompt, we optimize the generator with two kinds of CLIP-based guidance~\cite{gal2022stylegan}. However, leveraging text-guided 2D methods directly into the 3D setting is challenging as it tends to break 3D awareness and deformation awareness inherited in the generator parameters. To this end, we make some necessary modifications to the training framework. Please refer to the supplemental material for details. As shown in Fig.~\ref{inversion}, we achieve high-quality 3D-aware stylized portrait synthesis with preserving well properties (i.e. 3D consistency and accurate animation).
\section{Limitations and future work}

Though our approach enables reasonable extrapolation on some rare expressions (e.g. eye blinks, pouting, etc.), it struggles to model some other challenging expressions with full consistency, such as one-side mouth up, frown, sticking tongue out, etc.  We could use high-quality video clips with more abundant expressions for training as well as a more powerful face model for better extrapolation. We leave it for future work. Furthermore, our model has the potential to provide a strong 3D prior for accelerating person-specific avatar reconstruction. Besides, extending our methods into full-body settings is also a promising direction.
\section{Conclusion}
We have presented Next3D, a novel animatable 3D representation for unsupervised learning of high-quality and 3D-consistent virtual facial avatars from unstructured 2D images. Our approach has pushed the frontier of photorealistic animatbale 3D-aware image synthesis. Serving as a strong 3D prior, we believe our learned 3D representation will boost a series of downstream applications including 3D-aware one-shot facial avatars and animatable out-of-domain avatar generation. 

{\small
\bibliographystyle{ieee_fullname}
\bibliography{egbib}
}
\clearpage
\appendix
\section{Additional experiments}
\label{sec:1}

\subsection{Deformation-aware discriminator}
\begin{figure}[htbp]
  \centering
  \includegraphics[width=\linewidth]{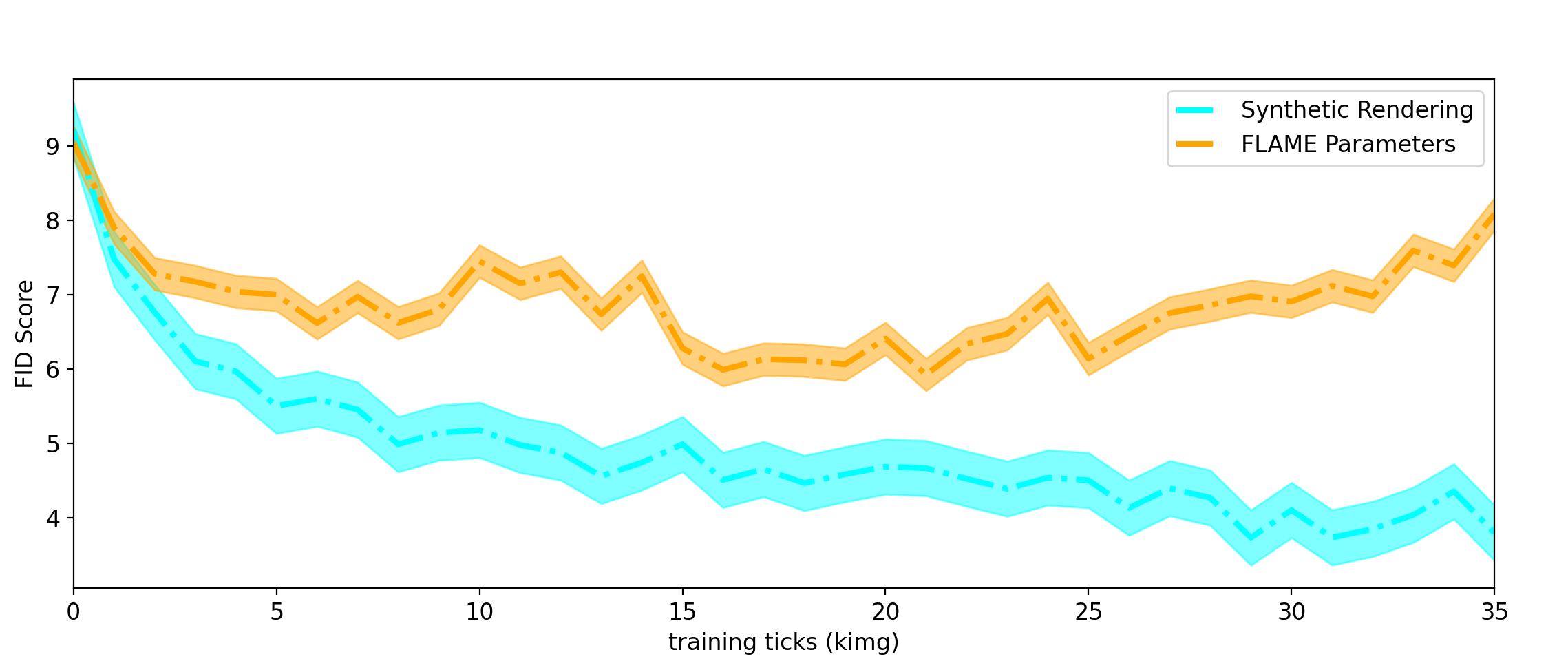}
  \caption{Training convergency with the discriminator designs.}
  \label{discriminator}
\end{figure}

We propose a deformation-aware discriminator which additionally takes the synthetic renderings as input. Furthermore, we also take experiments on the parameter conditioning method proposed in GNARF~\cite{bergman2022gnarf}. Specifically, we first train our model without either synthetic renderings or FLAME parameters conditioning for about two days. Then, we test two methods based on the same checkpoint and report the changing trend of FID scores for two methods in Fig.~\ref{discriminator}. The discriminator with synthetic rendering input converges to a better FID score, while the one conditioned on FLAME parameters incurs divergency. Note that we have added random noise to the FLAME parameters for better convergency following GNARF. 

\subsection{Training strategy of 3D-aware stylization}

\begin{figure}[htbp]
  \centering
  \includegraphics[width=\linewidth]{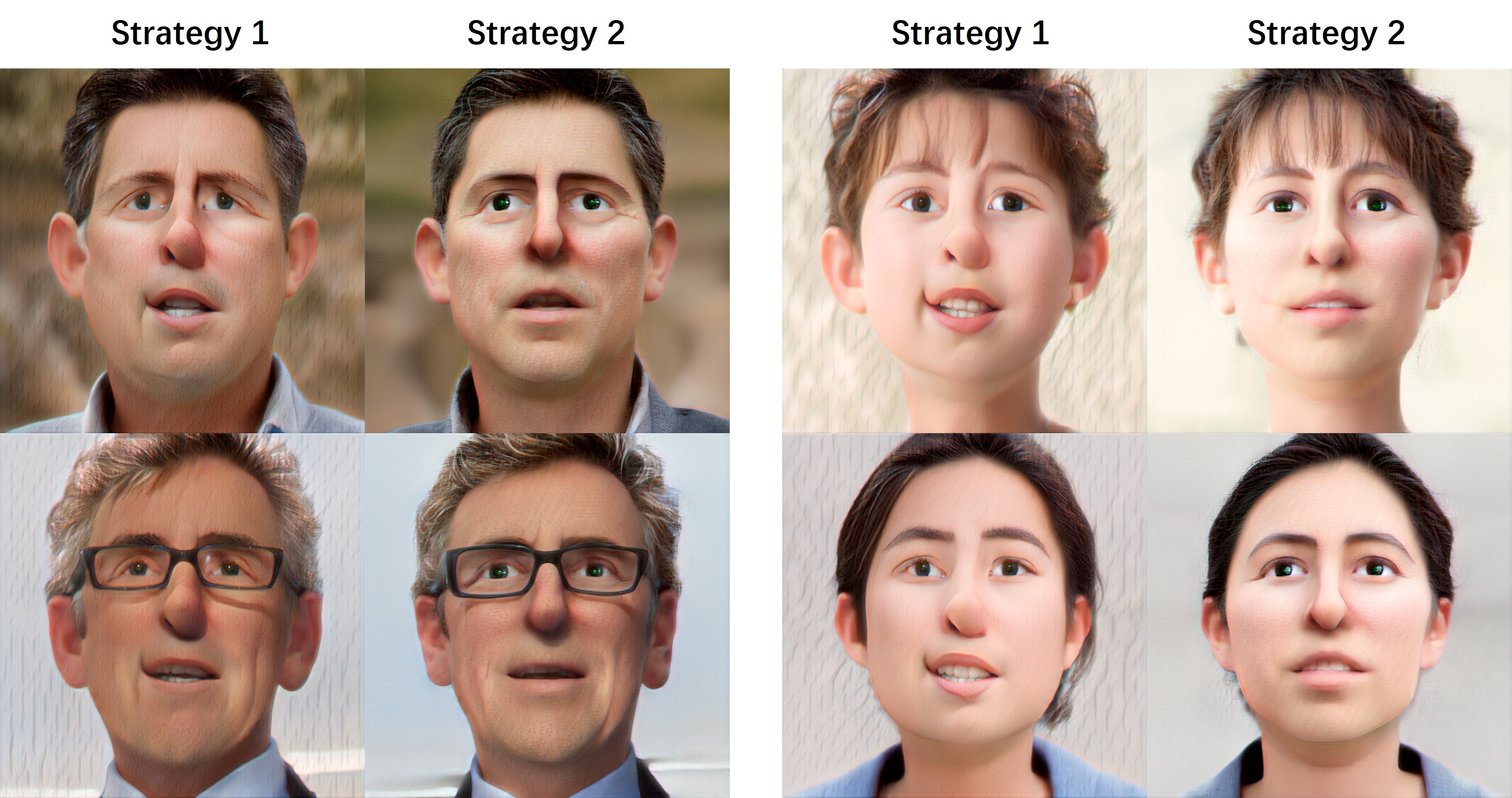}
  \caption{Ablation study on the training strategies of 3D-aware stylization.}
  \label{ablation_suppl}
\end{figure}

We conduct an ablation study on two strategies for freezing layers of the generator during 3D-aware stylization. The first one is the default setting following StyleGAN-NADA~\cite{gal2022stylegan} that freezes all toRGB layers in the synthesis network. Though it works in 2D space, we found it leads to degraded image quality and dissymmetry. To this end, we adopt another strategy which optimizes the last toRGB layer for each synthesis network. In our case, there are three StyleGAN-based synthesis network including a neural texture generator $G_{uv}$, a static tri-plane generator $G_{static}$, and a teeth completing module $G_{teeth}$ so we add the last toRGB layers of these three synthesis networks into optimization. As can be seen in  Fig.~\ref{ablation_suppl}, the second strategy improves the synthesis quality.
\section{Implementation details}
\label{sec:2}
We implemented our 3D GAN framework on top of the official PyTorch implementation of EG3D~\cite{eg3d}~\footnote{https://github.com/NVlabs/eg3d}. We adopt several hyperparameters and training strategies of EG3D including blurred real images at the beginning, pose-conditioned generator, density regularization, learning rates of the generator and discriminator. Due the limitation of computing material, we drop the two-stage training strategy and fix the neural rendering resolution to 64 and the final resolution to 512 instead. 

\subsection{Data preprocessing}

\begin{figure}[]
  \centering
  \includegraphics[width=0.8\linewidth]{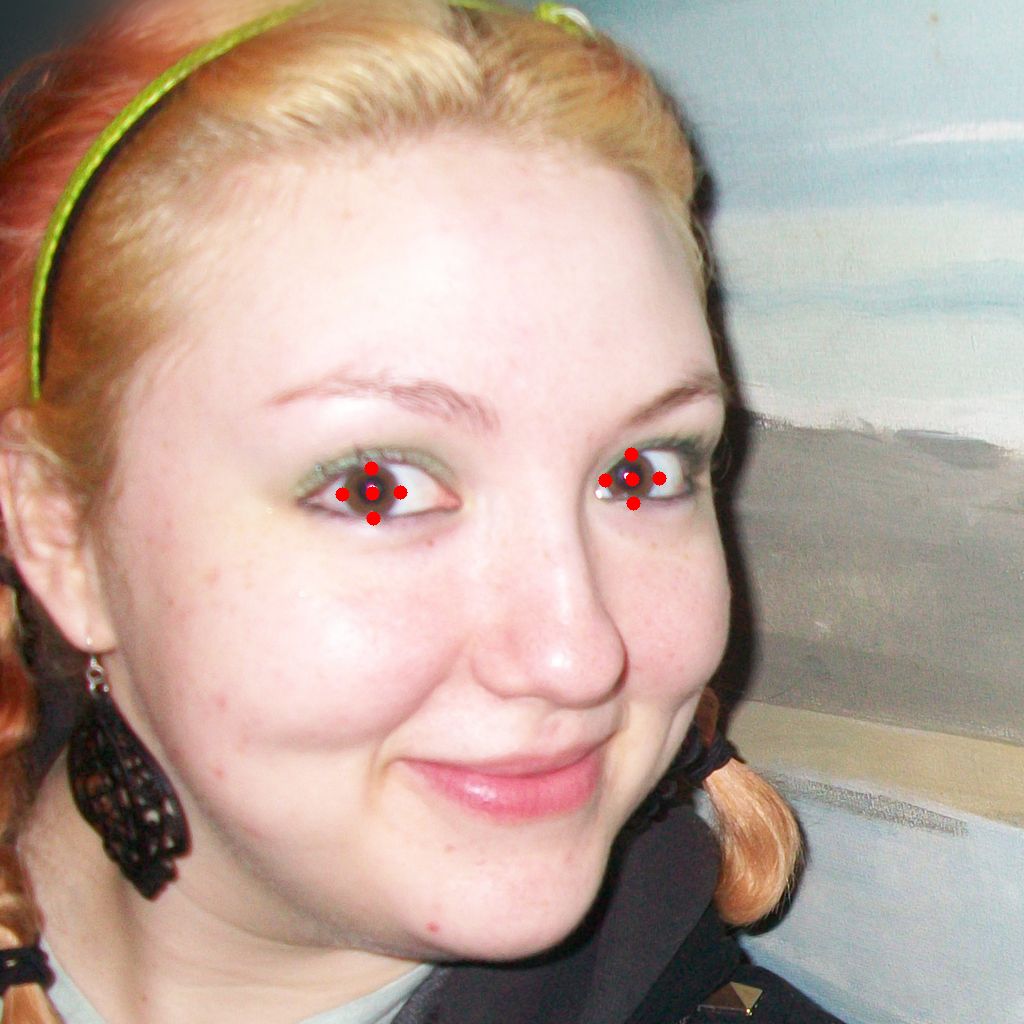}
  \caption{Detect the landmarks related to eyes.}
  \label{mediapipe}
\end{figure}

We use FLAME template model to drive the facial deformation and use DECA~\cite{feng2021learning} to extract FLAME parameters. Since there is no suitable model to accurately extract eye poses, we optimize eye poses with an off-the-shelf landmark detector~\footnote{https://mediapipe.dev/}. Specifically, the detector extracts five landmarks around the eyes, as shown in Fig.~\ref{mediapipe}. Accordingly, we select five vertices on the template mesh and the optimizable variables of eye poses are yaw and pitch. To optimize eye poses of a given portrait image, we minimize the re-projection errors of the vertices and detected landmarks by the PyTorch-implemented gradient descent. Since the FLAME template mesh has a different scale to the pretrained EG3D model, we initially rescale the template by 2.5 for a coarse visual alignment and fine-tune the translation and scale during training. 

\subsection{Generator}

\begin{figure}[]
  \centering
  \includegraphics[width=\linewidth]{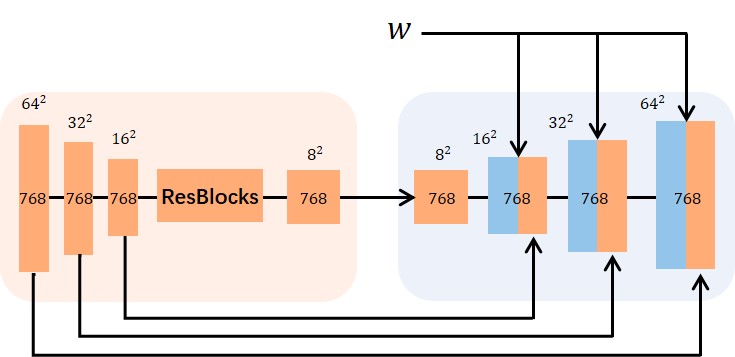}
  \caption{The detailed architecture of $G_{teeth}$.}
  \label{teeth}
\end{figure}

Our generator introduces a style-unet-based teeth completing module $G_{teeth}$ whose architecture is illustrated in Fig.~\ref{teeth}. The left part encodes the concatenated tri-plane teeth textures with dimensions of 768 (256 $\times$ 3) into multi-scale feature maps ranging from $64^2$ to $8^2$. Then the feature map with a resolution of $8^2$ is processed into the residual blocks and fed into the right generator as the input feature map. Finally, the generator outputs a $64 \times 64 \times 768$ feature map.

\section{Experiment details}
\label{sec:3}

\noindent\textbf{Inversion-based one-shot facial avatars.} We use an off-the-shelf face detector~\cite{deng2019accurate} to extract camera poses and crop the portraits in the wild to be consistent with the trainingset. We further extract the FLAME parameters and obtain the template mesh for each image by DECA~\cite{feng2021learning}. Following Pivotal Tuning Inversion (PTI)~\cite{roich2021pivotal}, we first optimize the latent code for 450 iterations and then fine-tune the generator weights for an additional 500 iterations. 

\noindent\textbf{3D-aware stylization.} Following StyleGAN-NADA~\cite{gal2022stylegan}, We optimize partial generator weights with others fixed.  In practice, we fixed all toRGB layers of the synthesis blocks except for the last ones for the texture generator and static generator. We also fix the NeRF decoders for preventing the 3D consistency from degeneration.

\section{Additional visual results}
\label{sec:4}

\begin{figure*}[htbp]
  \centering
  \includegraphics[width=\linewidth]{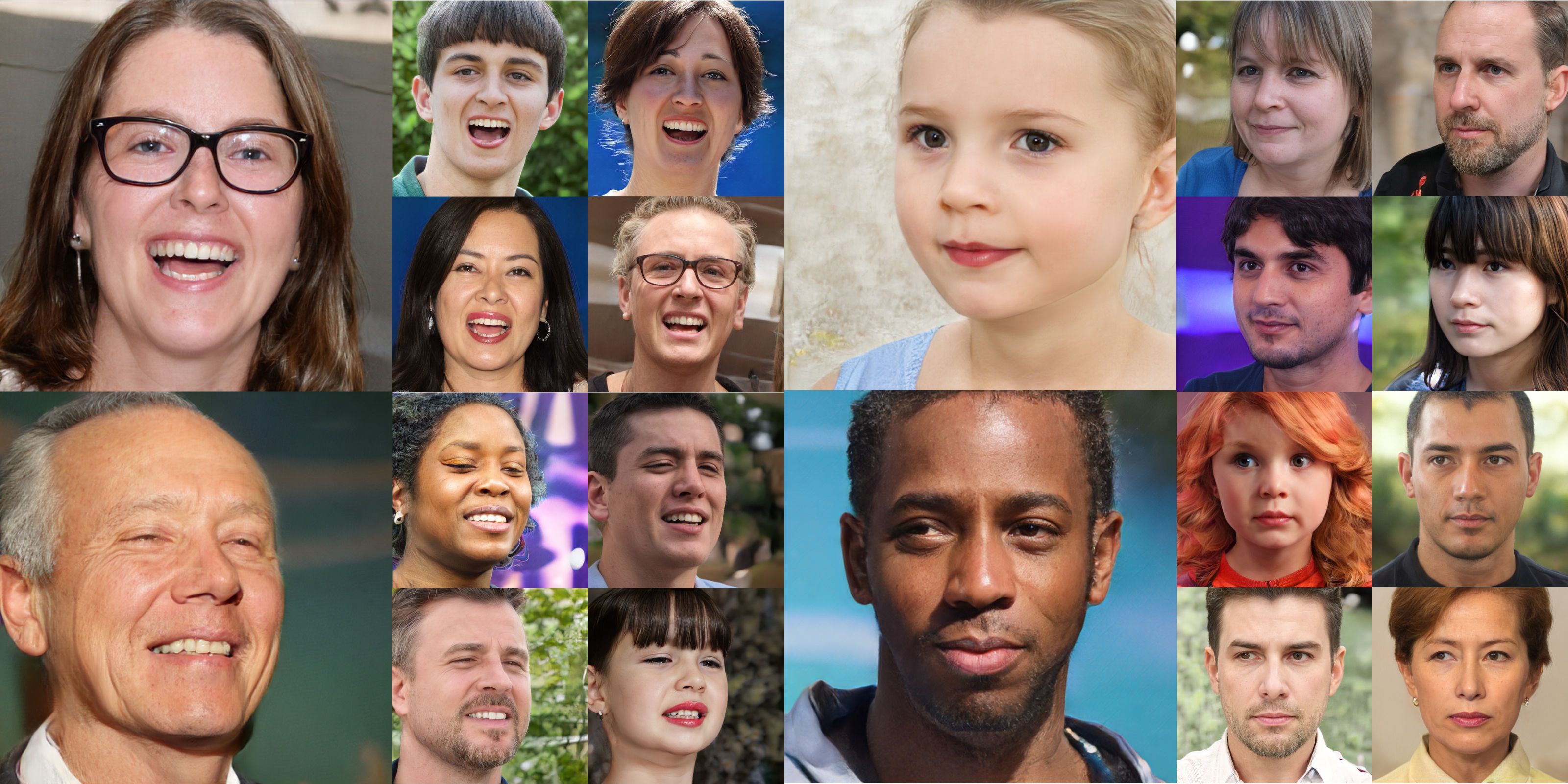}
  \caption{Generated examples with selected expressions and poses.}
  \label{examples_suppl}
\end{figure*}

\begin{figure*}[htbp]
  \centering
  \includegraphics[width=\linewidth]{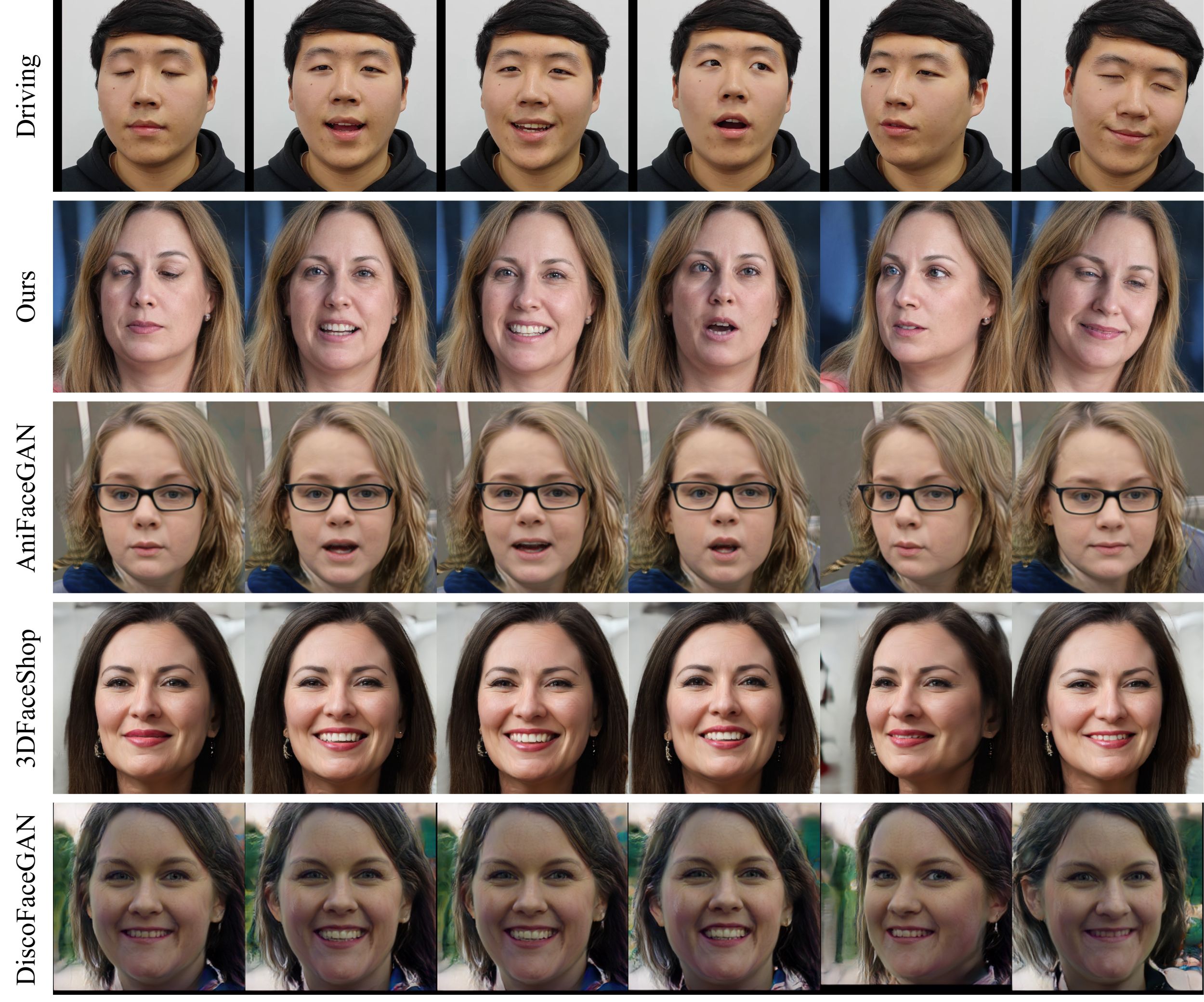}
  \caption{Qualitative comparison against baselines.}
  \label{comparison_suppl}
\end{figure*}

In this section, we provide additional visual results as a supplement to the main paper. Fig.~\ref{examples_suppl} provides selected examples of four certain expressions and poses, highlighting the image quality, expression controllability (e.g. gaze animation), and the diversity of outputs produced by our method. Fig.~\ref{comparison_suppl} provides a qualitative comparison against baselines on facial animation. 

\begin{figure*}[htbp]
  \centering
  \includegraphics[width=0.9\linewidth]{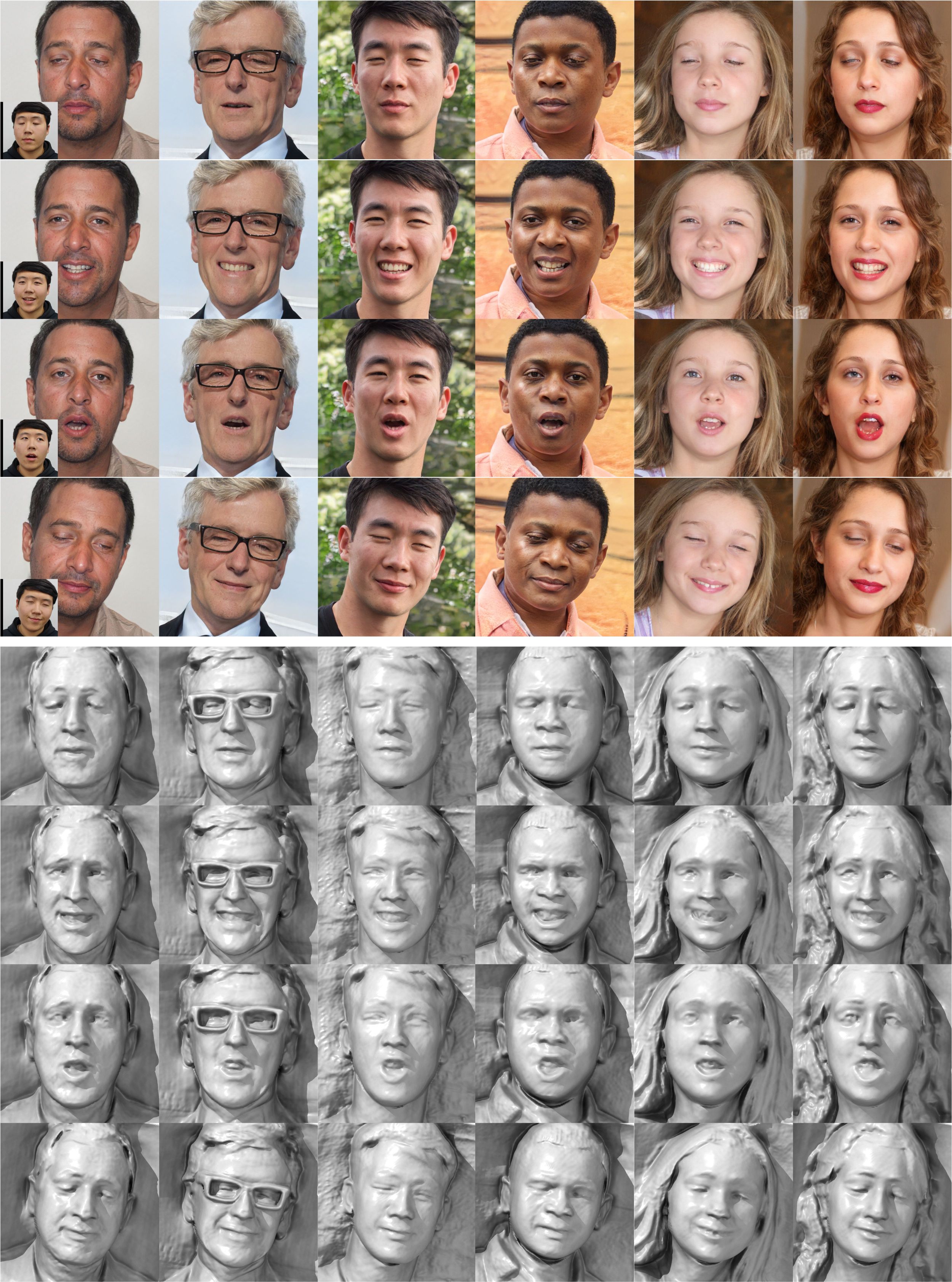}
  \caption{Animated virtual avatars with high-quality shapes.}
  \label{animation_suppl}
\end{figure*}

Fig.~\ref{animation_suppl} provides more results of animated virtual avatars with high-quality shapes. Note that the motions of eyelids can be reflected on the extracted meshes. Furthermore, the eyes are modeled as convex, suggesting that “hollow face illusion” is alleviatived. This is because while the gaze directions are highly pose-related, the rotated eyeballs in the template mesh provide an explicit gaze direction signal and thus helps to model such pose-related attribute and decouple them during inference.  

\begin{figure*}[htbp]
  \centering
  \includegraphics[width=0.95\linewidth]{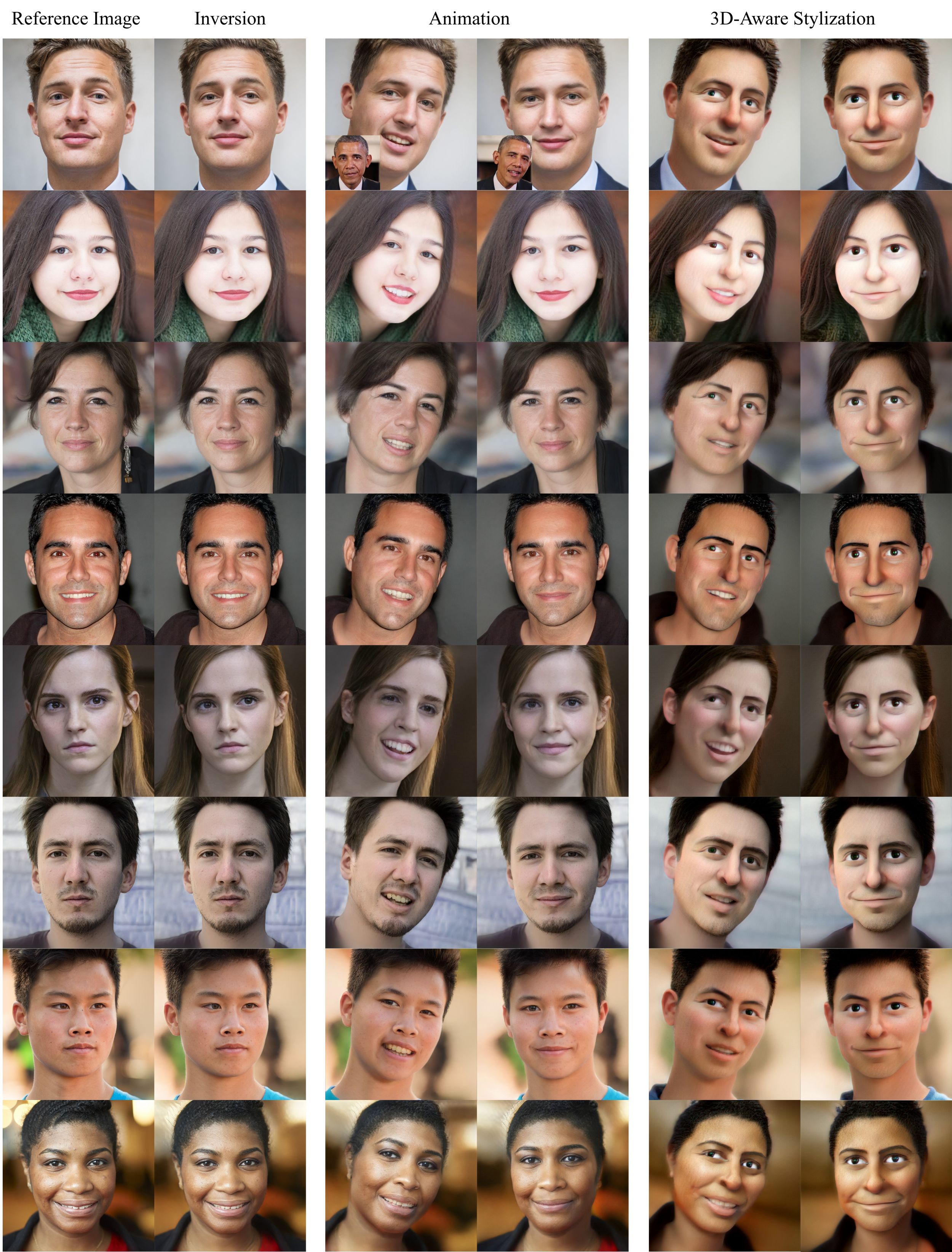}
  \caption{Visual results of one-shot avatars for real portraits and 3D-aware stylization.}
  \label{application_suppl}
\end{figure*}

Finally, we show additional results of the applications of our methods including one-shot avatars for real portraits and 3D-aware stylization in Fig.~\ref{application_suppl}. We encourage readers to view the accompanying supplemental video for the dynamic results. 

\clearpage
\clearpage

\clearpage

\clearpage

\clearpage

\end{document}